\relax
\documentclass[letterpaper]{article} 
\usepackage{aaai21}  
\usepackage{times}  
\usepackage{helvet} 
\usepackage{courier}  
\usepackage[hyphens]{url}  
\usepackage{graphicx} 
\usepackage{booktabs}
\usepackage{color}
\usepackage{xspace}
\usepackage{natbib}  
\usepackage{caption} 
\newcommand{\ourgym}{myGym\xspace}  

\title{\ourgym: Modular Toolkit for Visuomotor Robotic Tasks}
\author{Michal Vavrecka \textsuperscript{\rm 1}, Nikita Sokovnin \textsuperscript{\rm 1} \\ Megi Mejdrechova\textsuperscript{\rm 1} , Gabriela Sejnova\textsuperscript{\rm 1} , Marek Otahal\textsuperscript{\rm 1}}
\affiliations{

    \textsuperscript{\rm 1} Czech Institute of Informatics, Robotics and Cybernetics\\
    Czech Technical University in Prague, 
    Czech Republic

}
\begin{document}

\maketitle

\begin{abstract}
We introduce a novel virtual robotic toolkit \ourgym, developed for reinforcement learning (RL), intrinsic motivation, and imitation learning tasks trained in a 3D simulator. The trained tasks can then be easily transferred to real world robotic scenarios. The modular structure of the simulator enables users to train and validate their algorithms on a large number of scenarios with various robots, environments and tasks. Compared to existing toolkits (e.g. OpenAI Gym, Roboschool) which are suitable for classical RL, \ourgym \@ is also prepared for visuomotor (combining vision \& movement) unsupervised tasks that require intrinsic motivation, i.e. the robots are able to generate their own goals. There are also collaborative scenarios intended for human-robot interaction. The toolkit provides pretrained visual modules for visuomotoric tasks allowing rapid prototyping, and, moreover, users can customize the visual submodules and retrain with their own set of objects. In practice, the user selects the desired environment, robot, objects, task and type of reward as simulation parameters, and the training, visualization and testing themselves are handled automatically. The user can thus fully focus on development of the neural network architecture while controlling the behavior of the environment using predefined parameters. 
\end{abstract}

\section{Introduction}

\begin{figure}[h]
    \centering
    \includegraphics[width=.4\textwidth]{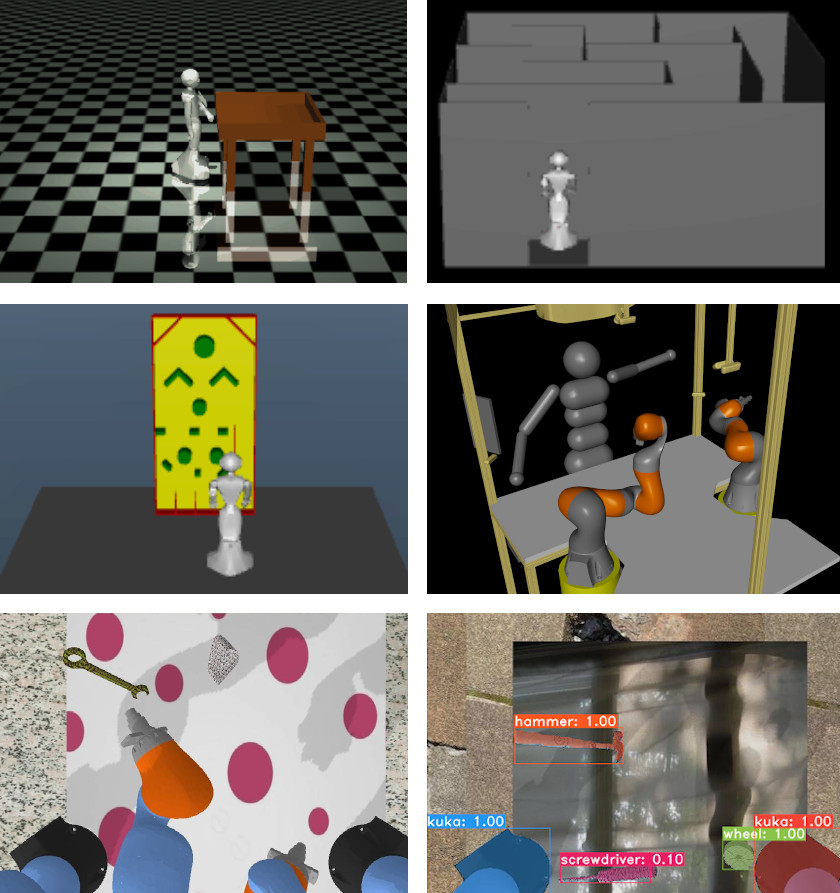}
    \caption{Examples of robotic workspaces. \textit{Top:} Pepper robot by a desk \textit{(left)} and in a maze \textit{(right)}. \textit{Middle:} A vertical plane workspace \textit{(left)}, and a collaborative workspace with two KUKA arms and a humanoid \textit{(right)}. \textit{Bottom:} Example of image augmentation during training, where random textures from a database are applied to the objects \textit{(left)}. A pretrained neural network Yolact can be used to preprocess the images \textit{(right)}.}
    \label{fig:augmentation}
\end{figure}

\begin{figure*}[tb] 
\centering
 \makebox[\textwidth]{\includegraphics[width=.8\paperwidth]{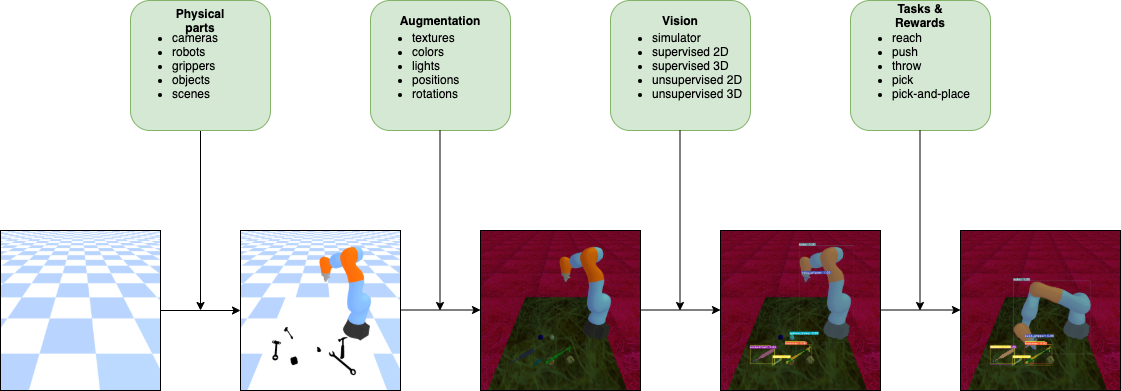}}
\caption{Overview of \ourgym \@ structure. The modular system helps to prepare desired environments and tasks to combine them compositionaly.The upper row stands for types of modules that adds specific features to the simulation. The bottom row standsfor the visualization of the environment after application of specific module.   }
\label{fig:gym_scheme}
\end{figure*}

With the growing complexity of recent neural network architectures, we  need to increase the amount of training data to provide good balance between neural network parameters and training instances. The most advanced neural networks nowadays consists of 175 billions of parameters \cite{brown2020language} and there is about 300 billions of instances provided during training. In case of textual data, there are big internet datasets available. In the area of visual datasets, the annotation of the data is manual and very time demanding. 

In the area of robotics, we require training data that are composed of visual inputs, actual positions of the objects in the environment, the current state of the robot actuators and also interactions within the environment. These type of datasets are the most expensive in terms of preparation and annotation. On the other hand, it is possible to generate an infinite amount of such training data in a simulator without any manual effort. There are virtual simulators that provide fully customizable environments and automatic annotation of all objects in the scene, serving as a robust source of training data. Since the recent simulators are also equipped with reliable physics engines and photorealistic rendering (see bellow), the neural architectures can be adopted in the real world tasks after successful training in the simulator.

We introduce novel toolkit called \textbf{\ourgym}, which serves as a modular platform for development and evaluation of cognitive architectures. Its main goal is to let the user fully focus on development and improvement of learning algorithms, while the simulation, task setup and visualization are automatically taken care of.

\section{Related work}
The possibility to train agents in a virtual environment has rapidly increased the progress in the area of reinforcement learning. This has naturally brought attention to further improvement of the simulation tools and, currently, one can select among many ready-to-use libraries and toolkits. Some of them are oriented on the gaming area \cite{beattie2016deepmind} \cite{bellemare2013arcade}, some on navigation tasks \cite{savva2017minos}\cite{savva2019habitat} or multi-agent systems \cite{song2019arena}. There is also large focus on a robot control and robotic manipulation and one of the widely used libraries in this area is OpenAI Gym \cite{brockman2016openai}. It is a collection of environments that includes algorithmic tasks, Atari and board games, several toy tasks, but also 2D and 3D robotic environments. There is a number of recent frameworks that extend OpenAI Gym (or are compatible with it) and here we provide their brief overview. For a detailed comparison, please refer to \cite{ivaldi2014tools}\cite{ferigo2020gym}.

An official extension of OpenAI Gym is their Robotic Environments toolkit (Roboschool), introduced along with a set of Baselines implementations of state-of-the-art RL algorithms like PPO or HER \cite{baselines}. However, it is based on the Mujoco physics engine \cite{todorov2012mujoco} which is not open source, similarly as the photorealistic virtual simulator from Nvidia ISAAC. 
An open source alternative is PyBullet \cite{coumans2016pybullet}, based on the widely used Bullet physics engine. PyBullet comes with its own set of environments (PyBullet-gym) with implementations of some basic agent-task scenarios. Gibson Env is also based on Bullet physics engine and provides a highly photorealistic environment for development of real-world perception in active agents \cite{xia2018gibson}. 

The two frameworks that are more closely related to \ourgym \@ toolkit are PyRoboLearn \cite{delhaisse2020pyrobolearn} and Gym-Ignition \cite{ferigo2020gym}. Gym-Ignition is designed for reproducible results and enables parallel or headless simulation, PyRoboLearn is highly modular and features a large number of robots and tasks including sports, locomotion, manipulation or control. However, neither of them has support for unsupervised tasks such as visuomotor manipulation or includes pretrained tools for visual processing. Moreover, most of the mentioned toolkits are discontinued (e.g. Roboschool) or their modular functions are not yet fully implemented (e.g PyRoboLearn). 

We would like to overcome the weak parts of the previous toolkits and extend the usecases to the area of intrinsic motivation and imitation learning by incorporating pretrained modules for vision. The user can then use them for both supervised and unsupervised learning. We consider the following features of our toolkit as the most advantageous:
\begin{itemize}
    \item interchangeable parts of the simulation
    \item parametric definition of the environment
    \item predefined intrinsic reward functions
    \item predefined task types
    \item pretrained neural networks for image preprocessing
    \item domain randomization
    \item visual augmentation
\end{itemize}

The highly modular structure of \ourgym \@ and its support for multiple physics engines (Mujoco, PyBullet) enables the user to easily set up a custom environment tailored to his needs, while the adjustable randomization and visual augmentation provide robust results that can be easily implemented in real world robotic scenarios.   

\begin{table*}[t]
  \caption{Comparison of features provided by different frameworks that are all based on / compatible with OpenAI Gym.}
  \label{comparison-table}
  \centering
  \begin{tabular}{|c|c|c|c|c|c|}
    \hline
    \textbf{Toolkit}     &  \textbf{Baselines} & \textbf{Pretrained Vision}     &     \textbf{Augmentation} & \textbf{Parametric env}  & \textbf{Parametric rewards}  \\
    \hline
    Roboschool & $\bullet$ & $\circ$ &  $\circ$ &  $\circ$  &  $\circ$  \\ 
    \hline
    PyBulletGym       & $\bullet$ &  $\circ$ &  $\circ$ &  $\circ$  &  $\circ$  \\
    \hline
    Gibson                      & $\bullet$ &  $\circ$ &  $\circ$ &  $\circ$ &  $\circ$   \\
    \hline
    PyRoboLearn                 & $\bullet$ &  $\circ$ & $\bullet$ & $\bullet$ &  $\circ$  \\
    \hline
    Gym-Ignition                &  $\circ$ &  $\circ$ &  $\circ$ &  $\circ$ &  $\circ$ \\
    \hline
    \textbf{\ourgym}          & $\bullet$ & $\bullet$ & $\bullet$ & $\bullet$ & $\bullet$ \\
    \hline
  \end{tabular}
\end{table*}


\section{Toolkit structure}

The  novel simulator toolkit - \ourgym \@ is intended for scientists interested in development of artificial neural network algorithms with application in robotic tasks. 
The user can choose among several workspaces, robots, manipulation tasks and levels of supervision. See Fig. \ref{fig:gym_scheme} for details.


\subsection{Environment}\label{AA}

Compared to previous toolkits that stem from OpenAI Gym structure, we decomposed the environment into a set of specific modules that can be combined together depending on user requirements. There are basic types of environments with basic arrangement that are suitable for manipulation tasks (worskpace, kitchen table), navigation tasks (maze, office room), curriculum learning (mini-golf, basketball or football playgrounds with increasing complexity) or collaborative and imitation learning (collaborative workspace or a kitchen unit). There are also basic methods to reset and control the environment and the connections to the physical engines that controls the simulation (Mujoco or PyBullet). The rest of the environment is modular based on the parameters defined by the user. Each environment in \ourgym \@ can be constructed from a few modules:
\begin{itemize}
\item Robot - represents a simulated robot and gives an interface to control it and get observation data
\item Objects - represents objects that the robot can manipulate
\item Camera - represents a camera with which images are taken
\item Task - represents a specific task to learn (reach, throw, pick-and-place, etc.)
\item Reward - represents the type of reward
\end{itemize}
Each environment also includes the interactive test mode, where the user can manipulate the objects and test the neural network effectiveness in a specific configuration. In the following paragraphs, we describe each module in greater detail.

\subsection{Robot and camera configuration}

\begin{figure}[h]
    \centering
    \includegraphics[width=.5\textwidth]{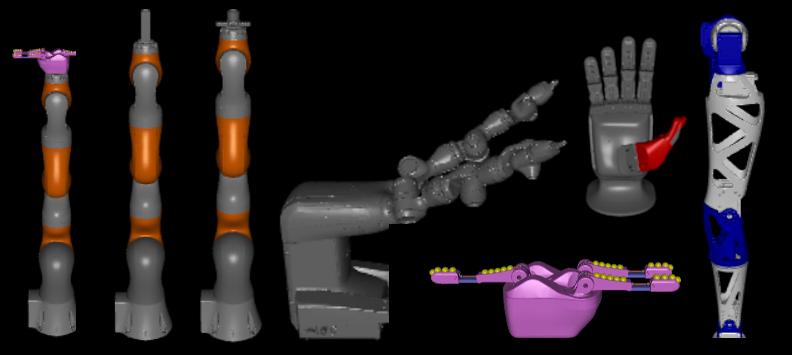}
    \caption{Examples of robotic arms and exchangeable grippers included in \ourgym. The user can easily add any robot to the environment just by specifying its name and optionally position, orientation and initial configuration.}
    \label{fig:robots}
\end{figure}

When the user chooses an environment with static objects, he can add the robot to the environment just by adding the parameter with specified name, position, orientation and initial configuration of the robot. The toolkit provides a set of industrial robots (Kuka IIWA, Franka Emika, Universal robots), research robots (Reachy arm, Gummi arm, Jaco arm), dual arm robot Yumi and also humanoid robot for navigation tasks (Pepper robot). The user can also connect a gripper that is suitable for the selected task as there are non-grasping grippers (magnetic gripper, Festo universal gripper), non-tactile grippers (Brunel Hand, Schunk two finger gripper) and also tactile grippers (RightHand Takktile and Barret Hand). You can see the examples of robots and grippers in Fig. \ref{fig:robots}. There are many combinations of robots and grippers, but the toolkit will not allow user to create an impossible configuration (e.g. to connect a tactile gripper to a humanoid robot). After a successful initialization of the the robot in the environment, the action space is automatically set according to the number of joints and tactile sensors.
The initialization of the cameras is similar to the initialization of the robot. In case that the robot is not equipped with visual sensors, the user can specify the amount of cameras, their positions and other properties in the configuration file. Otherwise, the cameras are fixed to the robot body (e.g. Pepper or Realsense at Franka Emika). The user then specify the active cameras that will be used during training and testing. 

\subsection{Manipulable objects configuration}

As most of the visuomotor tasks require object manipulation, the toolkit comes with several sets of objects such as household items, geometric primitives or industrial tools. These objects have been chosen as suitable for manipulation tasks and the available visual modules were pretrained on them. The user can also easily add custom objects to the toolkit and retrain the visual modules if needed.

The user defines a desired subset of all available objects during the configuration stage simply by stating their names. The user can further specify position in the scene, where each object will be initialized at the beginning of a training episode, by providing its spatial coordinates. It is also possible to randomize the initialization of objects by specifying an area instead of a position, varying the number of initialized objects and performing random choice of objects from the given subset.


\subsection{Task specification}

Compared to previous toolkits (e.g. Roboschool, PyRoboLearn, PyBullet-gym), we simplified the process of task specification and reward calculation. There is reward and task specified in each environment separately in the previous toolkits and the user has to manually modify the environment to a novel task. We improved this part of the simulator as there are generic types presented that should be easily adopted to any environment. We realize that most of the manipulation and navigation tasks share a common assumption about the reward calculation. At the beginning, the user has to identify objects in the environment that are crucial for reward calculation just by assigning them a target and a goal identifier. As most of the tasks (reach, push, throw, pick, place, navigate) requires just to calculate the distance between target and goal object (the object should be also part of robot body or position on the objects surface), this process can be easily set just by few parameters. As the user defines the type of the task and crucial objects, it is also necessary to define what source of information is used to calculate the position of objects. This process is described in the next section. 


\subsection{Reward calculation}

The specific training reward depends both on the selected type of task (i.e. push, pick and place, reach) and type of learning (supervised RL or unsupervised/intrinsic learning). The common objective for all rewards in our toolkit is minimization of distance between two instances - usually between two objects (pick and place, push tasks) or between the agent and an object (reaching task). 

In case of fully supervised learning, the reward is calculated based on ground truth positions of the agent and objects, as retrieved from the simulator. In case of the intrinsic motivation, the reward is calculated based on image input only - in such case, we provide pretrained neural submodules that can convert the image into a representation that can serve for reward calculation. For example, the distance between an initial and goal position of a single object can be calculated from latent representations encoded by a pretrained variational autoencoder. In case of two or more objects in the scene, the distance can be retrieved from centroids inferred by a pretrained YOLACT network. 

We also considered that the distance itself can be calculated in many different ways and thus influence the learning results. Therefore, we leave it up to the user to experiment and choose among several most commonly used distance functions (e.g. Euclidean, Manhattan or Mahalanobis distance). 

In practice, the user only needs to specify the task, the type of distance to minimize and whether to calculate the reward from ground truth or pre-processed image. It is also possible to exchange the pretrained visual networks for a custom architecture or to design an own, shaped reward to train on.


\subsection{Augmentation}

Data augmentation is a key component for simulation-to-reality (Sim2Real) transfer. In order to successfully apply reinforcement learning to real robotic tasks, the systems based on computer vision should be sufficiently accurate. To handle gaps between synthetic data and real-world images with an endless variety of material appearance, colors and lighting conditions, we use the simple and fast technique - domain randomization \cite{tobin2017domain}. The main purpose of this approach is to provide enough variability of simulated data at training time such that the model will be able to successfully generalize to real-world data during the test time.

The whole environment can be randomized at several levels: random colors/textures applied to all objects, random camera position and orientation, variable light position, type and intensity and several postprocessing filters. There is example of the virtual environment augmentation can be seen in Fig. \ref{fig:augmentation}.

Due to modularity, each environment based on PyBullet in \ourgym \@ can be easily randomized using a special wrapper. Randomization parameters can be managed by a configuration file. Each parameter is controlled by a corresponding randomizer component.

\begin{itemize}
\item Light Randomizer - randomizes the parameters of simulation lighting such as direction, color, distance, ambient, diffuse and specular. Desirable elements of light configuration and their ranges can be also specified.
\item Camera Randomizer - randomizes the position and orientation of the camera using for images rendering.
\item Texture Randomizer - randomizes objects texture using a specified path to folder with images.
\item Color Randomizer - randomizes color attributes of objects in the scene such as RGBA color and specular color of the object. It can be applied at the same time as Texture Randomizer.
\item Joint Randomizer - add random position to the robotic actuators
\item Postprocessing Randomizer - application of special filters to the camera image
\end{itemize}


\subsection{Pretrained modules}
The toolkit allows users to start building their applications in the shortest possible time by offering three already pre-trained modules helpful in visuomotor tasks. 

YOLACT \cite{bolya2019yolact} network performs fundamental image segmentation and object recognition. It is able to identify all robots and objects used in the toolkit and to determine their position in the scene from the camera inputs in real-time. DOPE \cite{tremblay2018deep} extends YOLACT capabilities allowing to obtain 6DoF information about objects - position and orientation in 3D space. VAE \cite{kingma2013auto} extracts important information about the scene by encoding the camera inputs from pixel to latent space. All three vision modules can be used for intrinsic motivation reward calculation, the VAE is even trained  an unsupervised manner.

In case the user wants to train a customized version of any of the vision modules, the toolkit provides an option to generate a training dataset. The user configures the environment, robots and objects in a very similar way as before training, and chooses what type and size of dataset to generate. At this stage, it is very helpful to use randomization that is also easily configurable. The training itself is further module-specific and the toolkit provides guidance on how to perform it.


\subsection{Training and testing}

The purpose of the RL training environment is to apply your 
model and continuously get the rewards from your defined utility function. 
A common entry point is a main script where you can 
specify the environment, robots, RL algorithm (e.g. baselines) or optional pretrained vision module. After training, it is possible to evaluate your model by running main script with test parameters, which will give you average final score. There is also an option to record your (visualized) run, which is useful for further sharing 
and analysis. There is a code for baseline training provided in form of oneliners with simulation parameters.

The Gym comes with pretrained submodules for vision, allowing to quickly 
start developing your RL model in a real life tasks. Without the actual 
need to train your models for vision, which is demanding on HW resources 
and a lengthy procedure. There are 2 main modes for vision - for unsupervised 
learning (using variational autoencoders), and supervised learning (YOLACT, DOPE).

\begin{figure}[h]
    \centering
    \includegraphics[width=.5\textwidth]{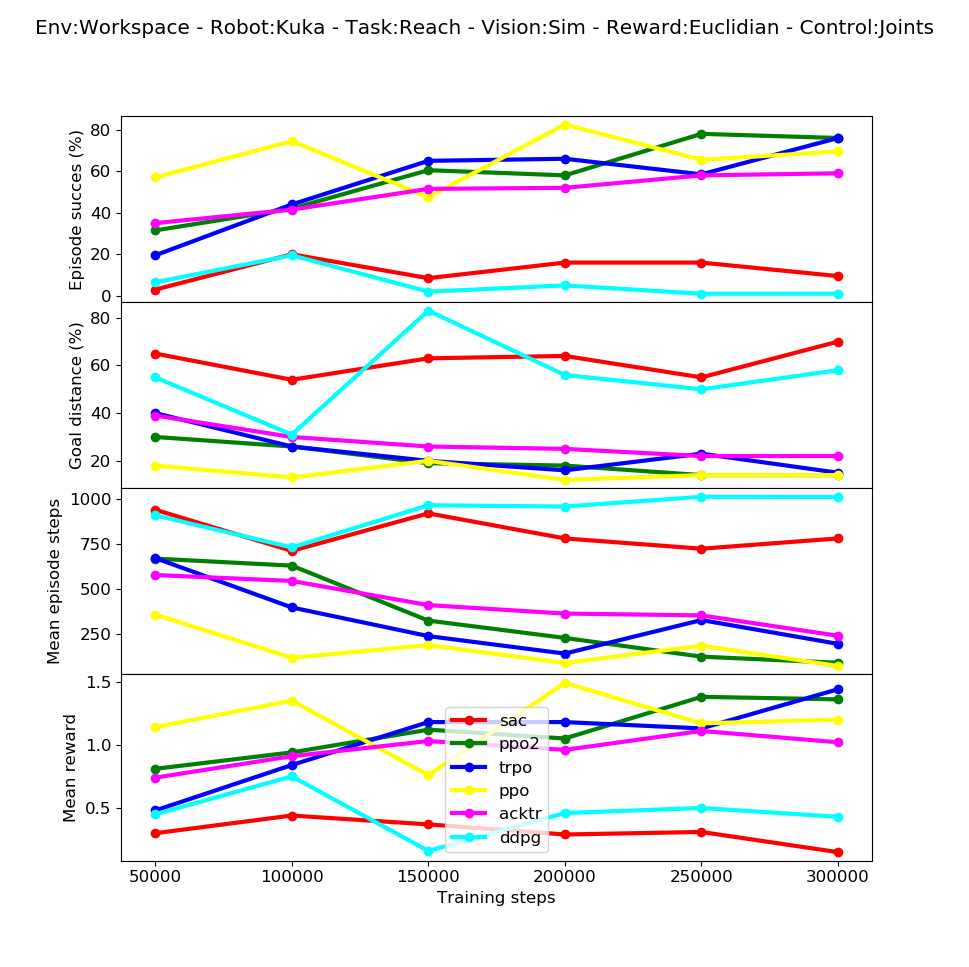}
    \caption{Evaluation protocol for direct comparison of the neural networks performance in the specific environment. The user can compare the performance of own neural network with the state of the art methods. }
    \label{fig:results}
\end{figure}

There are also functions for the visualization of the results after the training. As the researchers are interested in comparison of their own algorithms with respect to the state of the art, we provide evaluation protocol to directly compare the performance. You can see the example in Fig. \ref{fig:results}

\subsection {Sim2real - Transferring to physical sensors and robots with ROS}

As for many practical use-cases, interfacing with various hardware (sensors, cameras, robotic arms, etc.) is needed.  The industry golden-standard is Robot Operating System (ROS), which is a set of drivers and various libraries for designing robotic applications. 
Therefore our toolkit also has ROS2

The most common use-cases are twofolds: 
\begin{itemize}
    \item cameras (sensors): You can seamlessly switch (or combine) virtual cameras provided by the simulator environment, and any physical cameras. 
    \item robots (actuators): Since the common pitfalls of real HW equipment (expensive, slow, wear), users can experiment and train their architectures in the simulator. The provided models of robots are very precise, so the main idea is that it is possible to switch the entire pipe-line from simulator to your real environment. 
\end{itemize}

\subsubsection{Connecting pretrained network to real cameras and robots using ROS} 
The concept of ROS is a so called \textit{Node} (a physical HW or an algorithm), and a running application consists of a network of nodes that communicate (sending messages). 

As a part of the builtin functionality in our Gym, we provide a ROS2 node for the \textit{Intel RealSense camera}

We have extended our \textit{Vision submodule} to be used as a ROS node. This allows the user to train the vision system in simulator and then switch to ROS and use the trained model


\section{Conclusion}

We present a new virtual simulator toolkit \ourgym, which serves as a modular extension of the widely used OpenAI Gym simulator. Its main strengths lie in simple parametric definition of a custom composite environment, predefined task types, domain randomization and full visual augmentation. Furthermore, it is the first simulator toolkit with support for intrinsic motivation - this is achieved by predefined intrinsic reward functions and pretrained neural networks that can be used for image preprocessing and reward calculation instead of the ground truth. 

In our following work, we would like to further develop full support for all the mentioned features in both of the physics engines - currently, the closed nature of Mujoco physics engine does not enable some methods like visual augmentation or full randomization. 

We believe that wide variety of users will find the presented tool useful for their research. We have developed the toolkit so that the beginners can train neural networks in various environments while all necessary files are provided, and experts can modify it according to their needs.

\bibliography{paper}

\end{document}